\documentclass{ifacconf}

\makeatletter
\let\old@ssect\@ssect 
\makeatother

\usepackage{natbib}        
\usepackage{hyperref}

\makeatletter
\def\@ssect#1#2#3#4#5#6{%
  \NR@gettitle{#6}
  \old@ssect{#1}{#2}{#3}{#4}{#5}{#6}
}
\makeatother

\usepackage{url}
\usepackage{graphicx}      
\usepackage[export]{adjustbox}
\usepackage{comment}
\usepackage{color}
\usepackage{latexsym}
\usepackage{xurl}
\usepackage{mathtools}
\usepackage{caption}
\usepackage{subcaption}
\usepackage{amsmath,amssymb,amsfonts}
\usepackage{algorithmic}
\usepackage{pifont}
\usepackage{textcomp}
\usepackage{multirow}
\usepackage[table,xcdraw]{xcolor}
\usepackage{balance}

\DeclareMathAlphabet{\pazocal}{OMS}{zplm}{m}{n}

\begin{document}
\begin{frontmatter}

\title{Towards Sim2Real Transfer of Autonomy Algorithms using AutoDRIVE Ecosystem}

\author[First]{Chinmay Samak}
\author[First]{Tanmay Samak}
\author[First]{Venkat Krovi}

\address[First]{Department of Automotive Engineering, Clemson University International Center for Automotive Research, Greenville, SC 29607, USA. {\tt {\{\href{mailto:csamak@clemson.edu}{csamak}, \href{mailto:tsamak@clemson.edu}{tsamak}, \href{mailto:vkrovi@clemson.edu}{vkrovi}\}@clemson.edu}}}

\begin{abstract} 
The engineering community currently encounters significant challenges in the development of intelligent transportation algorithms that can be transferred from simulation to reality with minimal effort. This can be achieved by robustifying the algorithms using domain adaptation methods and/or by adopting cutting-edge tools that help support this objective seamlessly. This work presents AutoDRIVE, an openly accessible digital twin ecosystem designed to facilitate synergistic development, simulation and deployment of cyber-physical solutions pertaining to autonomous driving technology; and focuses on bridging the autonomy-oriented simulation-to-reality (sim2real) gap using the proposed ecosystem. In this paper, we extensively explore the modeling and simulation aspects of the ecosystem and substantiate its efficacy by demonstrating the successful transition of two candidate autonomy algorithms from simulation to reality to help support our claims: (i) autonomous parking using probabilistic robotics approach; (ii) behavioral cloning using deep imitation learning. The outcomes of these case studies further strengthen the credibility of AutoDRIVE as an invaluable tool for advancing the state-of-the-art in autonomous driving technology.
\end{abstract}


\end{frontmatter}


\section{Introduction}
\label{Section: Introduction}


The progression of connected autonomous vehicles (CAVs) necessitates a dual approach of cutting-edge research and comprehensive education. Research endeavors propel immediate advancements in the field, while education plays a pivotal role in equipping the next generation with the necessary skills and knowledge to propel the field even further in the long term. Recently, majority of researchers, educators and students rely on simulation tools and/or a scaled testbeds to to alleviate the monetary, spatial, temporal and safety constraints associated with rapid-prototyping of CAV solutions. In research settings, these platforms accelerate the process of designing experiments, recording datasets as well as re-iteratively prototyping and validating autonomy solutions. In educational contexts, such platforms facilitate the creation of interactive demonstrations, hands-on assignments, projects, and competitions centered around CAV technology.

\begin{figure}[t]
\centering
\includegraphics[width=\linewidth]{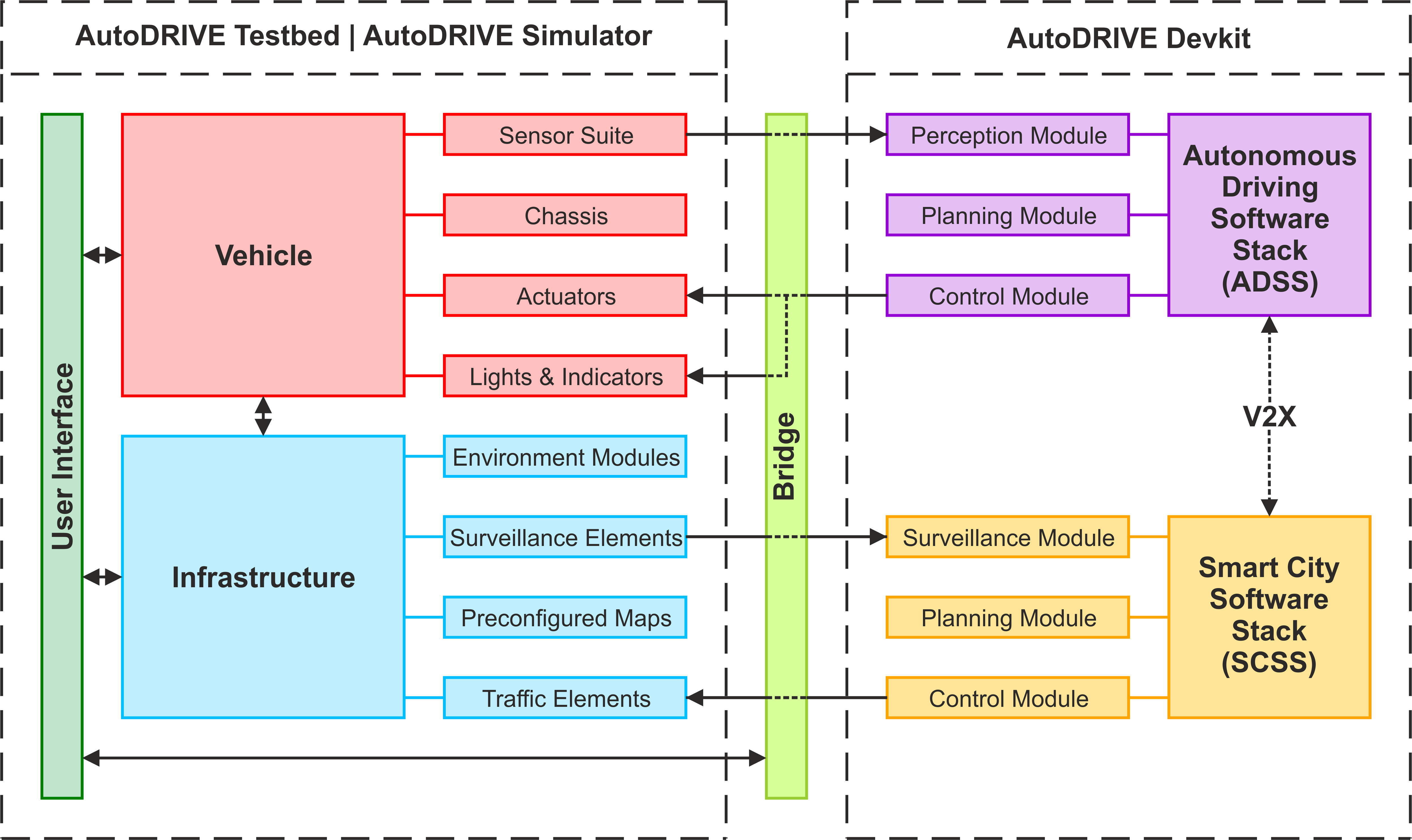}
\caption{AutoDRIVE Ecosystem offers of a tightly integrated trio for designing, simulating and deploying autonomy solutions using a unified workflow.}
\label{fig1}
\end{figure}


However, existing platforms catering to the development and validation of connected autonomy solutions have been observed to impose limitations on throughput. Firstly, a significant portion of these platforms lacks the essential integrity necessary to foster hardware-software co-development. Some platforms solely offer software simulators, while others merely provide scaled vehicles for testing autonomy algorithms. Such isolated platforms not only impede the prototyping phase due to compatibility issues but also hinder the validation phase involving the transition from simulation to real-world. Secondly, a majority of these platforms concentrate exclusively on vehicles, neglecting the holistic integration of an intelligent transportation ecosystem encompassing infrastructure, traffic elements, and peer agents. Consequently, their applications remain limited in scope. Thirdly, certain platforms are confined to specific domains or applications, featuring restricted sensing modalities, stringent design requirements, and fixed development frameworks. Such constraints further inhibit the versatility and adaptability of these platforms for broader use cases in connected autonomy research and education.

This research presents AutoDRIVE\footnote{\url{https://autodrive-ecosystem.github.io}} \cite{AutoDRIVEEcosystem2023}, a publicly accessible ecosystem specifically designed to facilitate the integrated development, simulation, and deployment of cyber-physical solutions pertaining to autonomous driving technology. This seamless workflow is made possible by a tightly integrated trio, consisting of an algorithm development kit for designing autonomy solutions, a software simulator for virtual prototyping and testing them under a variety of conditions and edge-cases, and a hardware testbed for physical deployment and validation (refer Fig. \ref{fig1}). The synergy among these three sub-systems not only enhances the hardware-software co-development of autonomy solutions but also effectively bridges the gap between simulation and reality. This work places particular emphasis on the challenges associated with bridging the sim2real gap for autonomy-oriented applications using the proposed ecosystem. In this context, we delve into the percepto-dynamics modeling and simulation aspects of a scaled vehicle and infrastructure using the proposed ecosystem. Furthermore, we substantiate our claims by showcasing the successful transition of two candidate autonomy algorithms from simulation to reality to help support our claims: (i) autonomous parking using probabilistic robotics approach for mapping, localization, path planning and control; (ii) behavioral cloning using computer vision and end-to-end deep imitation learning.


\section{State of the Art}
\label{Section: State of the Art}

\subsection{Software Simulators}
\label{Sub-Section: Software Simulators}

Over the years, open-source community has contributed several simulators for autonomous driving applications. Gazebo \cite{Gazebo2004}, natively integrated with Robot Operating System (ROS) \cite{ROS2009}, is commonly used for scaled autonomous robots. TORCS \cite{TORCS2021} has been an early focus in the self-driving community, particularly for manual and autonomous racing. Other notable examples include CARLA \cite{CARLA2017}, AirSim \cite{AirSim2018}, and Deepdrive \cite{Deepdrive2021}, developed using the Unreal 
game engine, as well as Apollo GameSim \cite{ApolloGameSim2021}, LGSVL Simulator \cite{LGSVLSimulator2020} and AutoRACE Simulator \cite{AutoRACE2021}, created using the Unity 
game engine.

However, these simulators pose 3 major limitations:

\begin{itemize}
    \item Firstly, some simulation tools prioritize photorealism over physical accuracy, while others prioritize physical accuracy over graphics quality. In contrast, AutoDRIVE Simulator strikes a balance between physics and graphics, providing a range of configurations to accommodate varying compute capabilities.
    \item Secondly, the dynamics and perception systems of scaled vehicles and environments differ significantly from their full-scale counterparts. Existing simulation tools do not adequately support scaled ecosystems to their full capacity. Consequently, transitioning from full-scale simulation to scaled real-world deployment necessitates substantial additional effort to re-tune the autonomy algorithms.
    \item Thirdly, the existing simulators may lack precise real-world counterparts, rendering them unsuitable for ``digital twin'' applications, involving synthetic data generation, variability testing, reinforcement learning, real2sim and sim2real transfer, among others.
\end{itemize}



\subsection{Hardware Testbeds}
\label{Sub-Section: Hardware Testbeds}


In recent times, numerous educational institutions have embarked on the development of scaled autonomous vehicles. Popular examples include the MIT Racecar \cite{MIT-Racecar2017}, F1TENTH \cite{F1TENTH2019}, and AutoRally \cite{AutoRally2021}. Additionally, community-driven platforms like HyphaROS RaceCar \cite{HyphaROS-Racecar2021} and Donkey Car \cite{DonkeyCar2021} have emerged, tailored to specific applications like map-based navigation and vision-aided imitation learning, respectively.
Duckietown \cite{Duckietown2017} is another platform, which employs differential-drive robots instead of kinodynamically constrained car-like vehicles, thereby falling short of the community's requirements. Nevertheless, it remains a popular platform for teaching autonomy fundamentals, much like TurtleBot3 \cite{Turtlebot2021}.

However, these platforms pose 3 major limitations:

\begin{itemize}
    \item Firstly, some of these platforms lack diverse sensing modalities, sufficient computational power, constrained actuation mechanisms, and active or passive infrastructural elements.
    \item Secondly, most platforms utilize commercial radio-controlled (RC) cars as their base-chassis, which are expensive, may not be readily available worldwide, and limit exploration in the realm of mechatronics engineering. Additionally, many platforms are confined to specific software frameworks like ROS, creating a skill-set dependency for end-users.
    \item Thirdly, some platforms lack any form of simulation support, some support simulation using RViz \cite{RViz2021} or Gazebo, while others provide task-specific Gym \cite{OpenAIGym2016} environments for machine learning; none of which is ideal.
\end{itemize}

\section{AutoDRIVE Ecosystem}
\label{Section: AutoDRIVE Ecosystem}

This section primarily focuses on digital-twin capabilities of AutoDRIVE Ecosystem, highlighting the strong resemblance between AutoDRIVE Simulator \cite{AutoDRIVESimulator2021, AutoDRIVESimulatorReport2020} and AutoDRIVE Testbed for seamless sim2real transfer of autonomy algorithms developed using AutoDRIVE Devkit.


\subsection{Physical Vehicle}
\label{Sub-Section: Physical  Vehicle}

\begin{figure}[htpb]
\centering
\includegraphics[width=\linewidth]{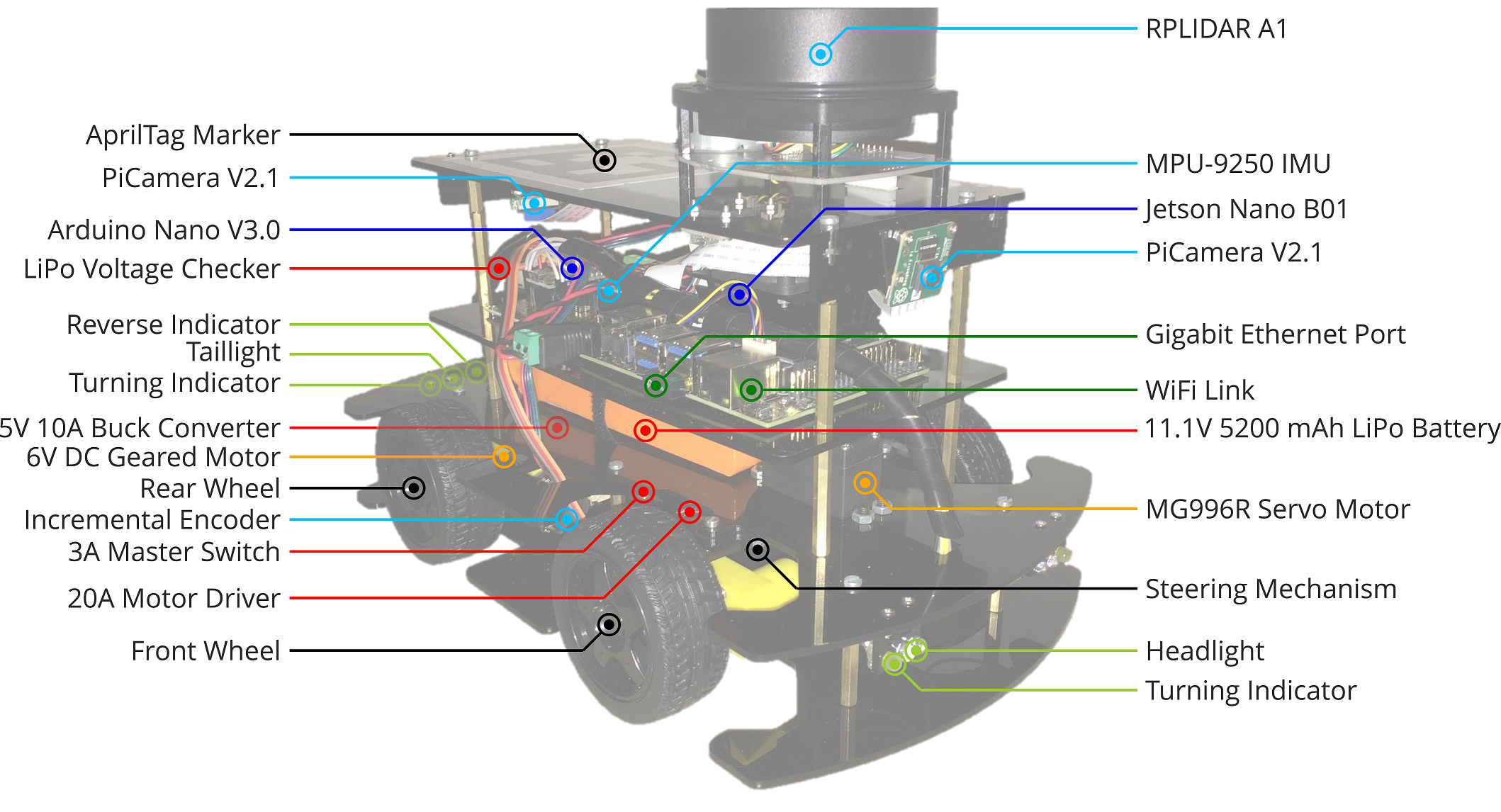}
\caption{Native vehicle (Nigel) of AutoDRIVE Ecosystem with its components and sub-systems highlighted.}
\label{fig2}
\end{figure}

AutoDRIVE's native vehicle, named Nigel (refer Fig. \ref{fig2}), offers realistic driving and steering actuation, comprehensive sensor suite, high-performance computational resources, and a standard vehicular lighting system.

\subsubsection{Chassis}
\label{Sub-Sub-Section: Chassis}

Nigel is a 1:14 scale vehicle, which adopts front/rear/all-wheel-drive Ackermann-steered mechanism, thereby resembling car-like kinodynamic constraints.

\subsubsection{Power Electronics}
\label{Sub-Sub-Section: Power Electronics}

An 11.1 V 5200 mAh lithium-polymer (LiPo) battery acts as the power-source for the vehicle. Other components such as the buck converter, motor driver, switch and voltage checker help route the raw power to appropriate sub-systems of the vehicle.

\subsubsection{Sensor Suite}
\label{Sub-Sub-Section: Sensor Suite}

Nigel hosts a comprehensive sensor suite comprising throttle and steering feedbacks, 1920 CPR incremental encoders, 3-axis IPS, 9-axis IMU, two 62.2$^\circ$ FOV cameras with 3.04 mm focal length and a 7-10 Hz, 360$^\circ$ FOV LIDAR with 0.15-12 m range and 1$^\circ$ resolution.

\subsubsection{Computation, Communication and Software}
\label{Sub-Sub-Section: Computation, Communication and Software}


Nigel relies on Jetson Nano Developer Kit - B01 for high-level computation (autonomy algorithms), communication (V2V and V2I), and software installation (JetPack SDK and AutoDRIVE Devkit). It also utilizes Arduino Nano (running the vehicle firmware) for sensor data acquisition and filtering, and actuators/lights control.

\subsubsection{Actuators}
\label{Sub-Sub-Section: Actuators}

Nigel is equipped with two 6V 160 RPM rated DC geared motors to drive its rear wheels, and a 9.4 kgf.cm servo motor (saturated at $\pm$ 30$^\circ$ w.r.t. zero-steer value) to steer its front wheels. All the actuators are operated at 5V, providing a top speed of $\sim$0.26 m/s for driving and $\sim$0.42 rad/s for steering.

\subsubsection{Lights and Indicators}
\label{Sub-Sub-Section: Lights and Indicators}

Nigel's lighting system comprises of dual-mode headlights, triple-mode turning indicators, and automated taillights and reverse indicators.

\subsection{Virtual Vehicle}
\label{Sub-Section: Virtual Vehicle}

\begin{figure}[htpb]
\centering
\includegraphics[width=\linewidth]{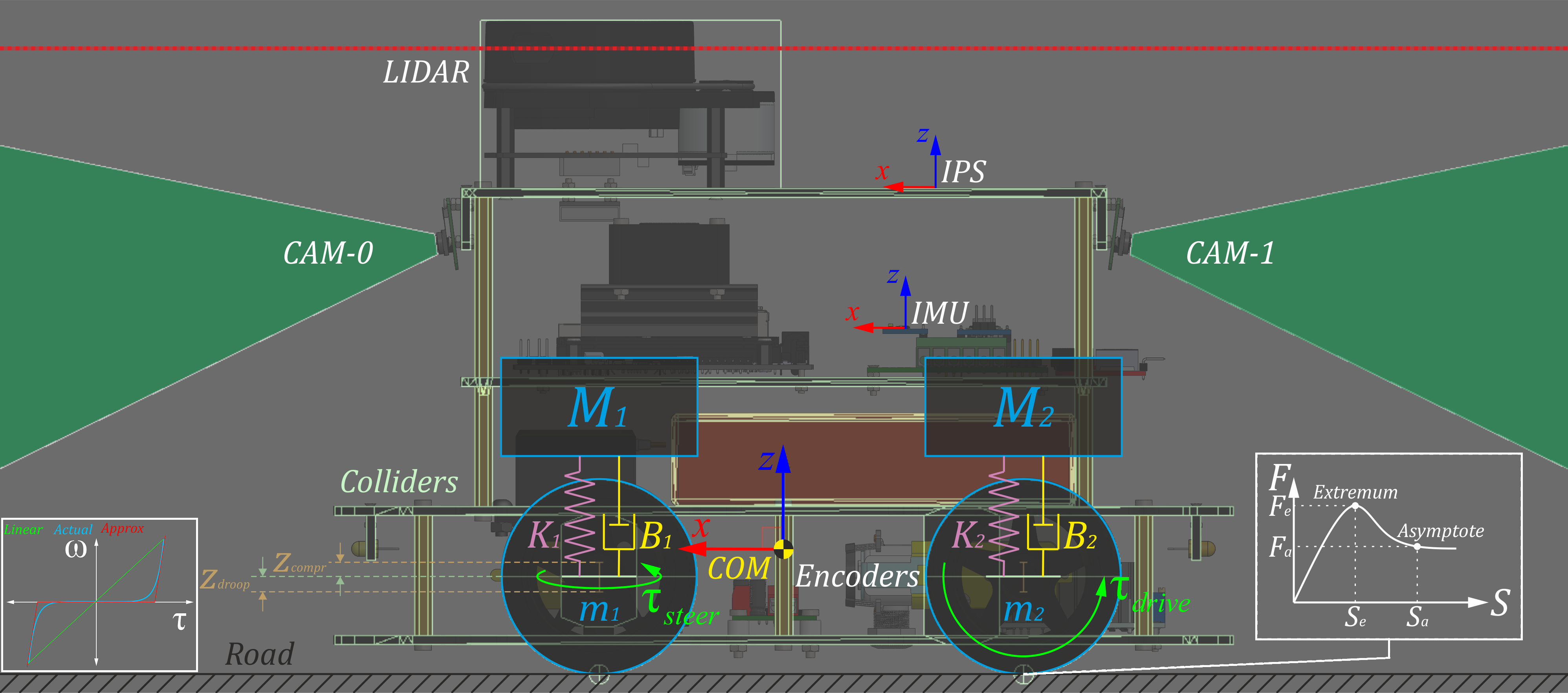}
\caption{Simulation of vehicle dynamics, sensors and actuators. The left inset depicts actuator dynamics model and the right inset depicts tire dynamics model.}
\label{fig3}
\end{figure}

The vehicle is jointly modelled as a rigid-body and a collection of sprung masses with inherent damping (refer Fig. \ref{fig3}). The ``sprung-mass'' representation computes the suspension forces, which, aggregated with the tire forces, are applied to the ``rigid-body'' representation that exactly mimics the mass, center of mass and moment of inertia of the ``sprung-mass'' representation. Needless to say, the two representations are related through the rigid-body center of mass equation: $X_{COM} = \frac{\sum{{^iM}*{^iX}}}{\sum{^iM}}$; where, $X_{COM}$ is the rigid-body center of mass offset, $^iM$ are the sprung masses such that $M=\sum{^iM}$ is the total mass of rigid-body, and $^iX$ are the sprung mass coordinates w.r.t. $^iM$.

\subsubsection{Suspension Dynamics}
\label{Sub-Sub-Section: Suspension Dynamics}

The vehicle is modeled with a rather stiff suspension system to simulate the selective passive compliance between wheel mounts and vehicle chassis to account for losses at these interfaces due to vibration, friction, wear, damping, loosening, deformation, fatigue and fretting. The suspension force acting on each of the sprung masses can be then computed using a second-order dynamic model: ${^iM} * {^i{\ddot{Z}}} + {^iB} * ({^i{\dot{Z}}}-{^i{\dot{z}}}) + {^iK} * ({^i{Z}}-{^i{z}})$; where, $^iB$ and $^iK$ are the damping and spring coefficients of $i$-th suspension, respectively. The computed suspension forces jointly affect the rigid-body dynamics of the vehicle as well as the tire forces being computed at that time instant (since it affects the load bearing down on the tires).

\subsubsection{Wheel Dynamics}
\label{Sub-Sub-Section: Wheel Dynamics}

The wheels of the vehicle are also modelled as rigid bodies of mass $m$, acted upon by gravitational and suspension forces: ${^im} * {^i{\ddot{z}}} + {^iB} * ({^i{\dot{z}}}-{^i{\dot{Z}}}) + {^iK} * ({^i{z}}-{^i{Z}})$; and the wheel rotations are damped to mimic rotational losses due to rolling resistance.

\subsubsection{Tire Dynamics}
\label{Sub-Sub-Section: Tire Dynamics}

The tire forces are broken down into longitudinal $F_{t_x}$ and lateral  $F_{t_y}$ components, and are computed based on the respective friction curve for each tire: $\left\{\begin{matrix} {^iF_{t_x}} = F(^iS_x) \\{^iF_{t_y}} = F(^iS_y) \\ \end{matrix}\right.$; where, $^iS_x$ and $^iS_y$  are the longitudinal and lateral slips of $i$-th tire, respectively. Here, the friction curve is approximated as a two-piece spline $F(S)$; one from zero $(S_0,F_0)$ to extremum point $(S_e,F_e)$, and other from extremum point $(S_e,F_e)$ to asymptote point $(S_a,F_a)$ (refer right inset in Fig. \ref{fig3}): $F(S) = \left\{\begin{matrix} f_0(S); \;\; S_0 \leq S < S_e \\ f_1(S); \;\; S_e \leq S < S_a \\ \end{matrix}\right.$; where, $f_k(S) = a_k*S^3+b_k*S^2+c_k*S+d_k$ is a cubic polynomial function, and $F(S)$ is saturated after the asymptote point $(S_a,F_a)$.

Now, slip is in-turn dependent on the various factors like tire stiffness, steering angle, wheel speeds, suspension forces and rigid-body momentum. Longitudinal slip $S_x$ is determined based on the difference between the longitudinal components of surface velocity of the wheel compared to the angular velocity of the wheel: ${^iS_x} = \frac{{^ir}*{^i\omega}-v_x}{v_x}$; where, $v_x$ is the longitudinal linear velocity of the vehicle (i.e., surface velocity of the wheel), $^ir$ is the radius of $i$-th wheel, and $^i\omega$ is the angular velocity of $i$-th wheel. Lateral slip $S_y$ is determined by the angle (commonly denoted as $\alpha$) between the direction the tire is moving in and the direction the tire is pointing in: ${^iS_y} = \frac{v_y}{\left| v_x \right|}$; where, $v_x$ is the longitudinal linear velocity of the vehicle, and $v_y$ is the lateral linear velocity of the vehicle (a.k.a. sideslip velocity).	

\subsubsection{Sensor Simulation}
\label{Sub-Sub-Section: Sensor Simulation}

As described earlier, the vehicle is provided with an abundance of sensing modalities, all of which are modeled and simulated close to their real-world counterparts.

\begin{itemize}
    \item \textit{Throttle Feedback:} Instantaneous feedback of throttle command $[-1,1]$, where positive values indicate driving forward and negative values indicate driving reverse.
    \item \textit{Steering Feedback:} Instantaneous feedback of steering command $[-1,1]$, where positive values indicate left turns and negative values indicate right turns.
    \item \textit{Incremental Encoders:} These are simulated by measuring the rotation of each of the rear wheels (based on their rigid-body transform update) and factoring in the resolution of the encoders.
    \item \textit{IPS:} Position of the vehicle is measured based on its rigid-body transform update. The values are converted from Unity's left-handed coordinate system to the right-handed coordinate system widely adopted for robotics applications. This mimics the AprilTag-based fiducial localization system on the physical vehicle.
    \item \textit{IMU:} Orientation of the vehicle is measured based on its rigid-body transform update. Additionally, the linear acceleration and angular velocity of the vehicle are measured based on temporally-coherent rigid-body transformations, using rigid-body equations of motion. This mimics the MPU-9250 on the physical vehicle.
    \item \textit{LIDAR:} Planar laser scan is recorded by 360$^\circ$ iterative ray-casting at 1$^\circ$ resolution and 7 Hz update rate. The raycast hits are recorded between the minimum (0.15 m) and maximum (12.0 m) ranges of the LIDAR, respectively. This mimics the RPLIDAR A1 on the physical vehicle.
    \item \textit{Cameras:} Physical cameras are simulated based on their focal length (3.04 mm), field of view (62.2$^\circ$), sensor size (4.6 mm) and target resolution (720p). Additionally, lens and film effects are simulated by post-processing the raw frames. This mimics the two PiCamera V2.1 modules on the physical vehicle.
\end{itemize}

\subsubsection{Actuator Simulation}
\label{Sub-Sub-Section: Actuator Simulation}

As described earlier, the vehicle has driving and steering actuators, the response delays and saturation limits of which are matched with their real-world counterparts by tuning their torque profiles and actuation limits, respectively (refer left inset in Fig. \ref{fig3}).

\begin{itemize}
    \item \textit{Driving Actuators:} Each of the rear wheels is driven using a rotary motor, which applies a torque to it: ${^i\tau_{drive}} = {^iI_w}*{^i\dot{\omega}_w}$; where, ${^iI_w} = \frac{1}{2}*{^im_w}*{^i{r_w}^2}$ is the moment of inertia of $i$-th wheel, and $^i\dot{\omega}_w$ is the angular acceleration of $i$-th wheel. Additionally, the holding torque of the driving actuators is simulated by applying an idle motor torque equivalent to the braking torque: ${^i\tau_{idle}} = {^i\tau_{brake}}$.
    \item \textit{Steering Actuators:} The front wheels are steered using a steering actuator coupled to the steering mechanism of the vehicle. The steering actuator also produces a torque proportional to the moment of inertia of the steering mechanism: $\tau_{steer} = I_{steer}*\dot{\omega}_{steer}$. The individual turning angles, $\delta_l$ and $\delta_r$, for left and right wheels, respectively, are calculated based on the commanded steering angle $\delta$, using the Ackermann steering geometry defined by wheelbase $l$ and track width $w$, as follows: $\left\{\begin{matrix} \delta_l = \textup{tan}^{-1}\left(\frac{2*l*\textup{tan}(\delta)}{2*l+w*\textup{tan}(\delta)}\right) \\ \delta_r = \textup{tan}^{-1}\left(\frac{2*l*\textup{tan}(\delta)}{2*l-w*\textup{tan}(\delta)}\right) \\ \end{matrix}\right.$
\end{itemize}

\subsection{Physical Infrastructure}
\label{Sub-Section: Physical Infrastructure}

AutoDRIVE provides a modular and reconfigurable infrastructure development kit, enabling swift design and construction of customized driving scenarios.

\begin{figure}[htpb]
     \centering
     \begin{subfigure}[b]{0.49\linewidth}
         \centering
         \includegraphics[width=\linewidth]{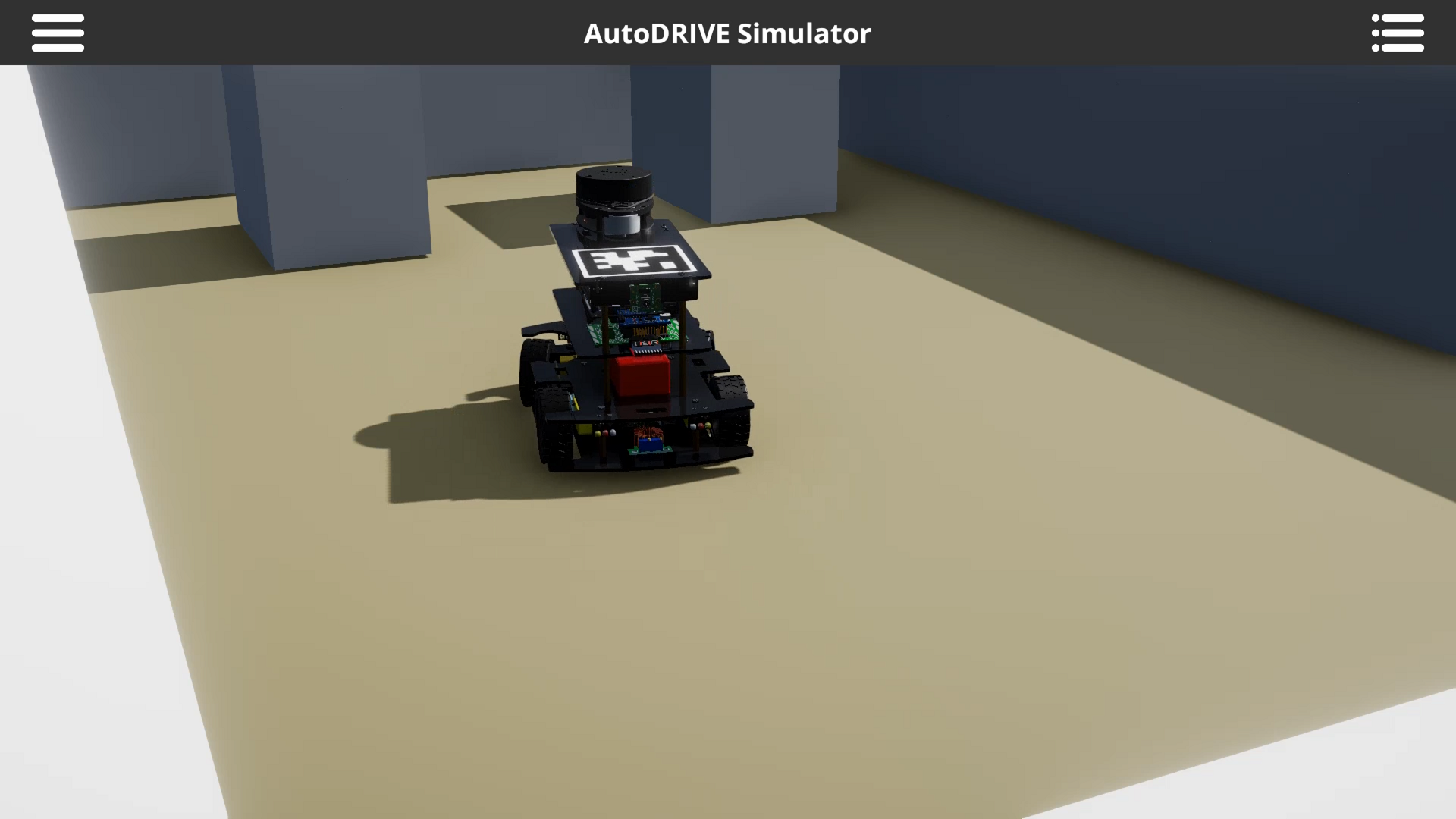}
         \caption{Parking School in simulation.}
         \label{fig4a}
     \end{subfigure}
     \hfill
     \begin{subfigure}[b]{0.49\linewidth}
         \centering
         \includegraphics[width=\linewidth]{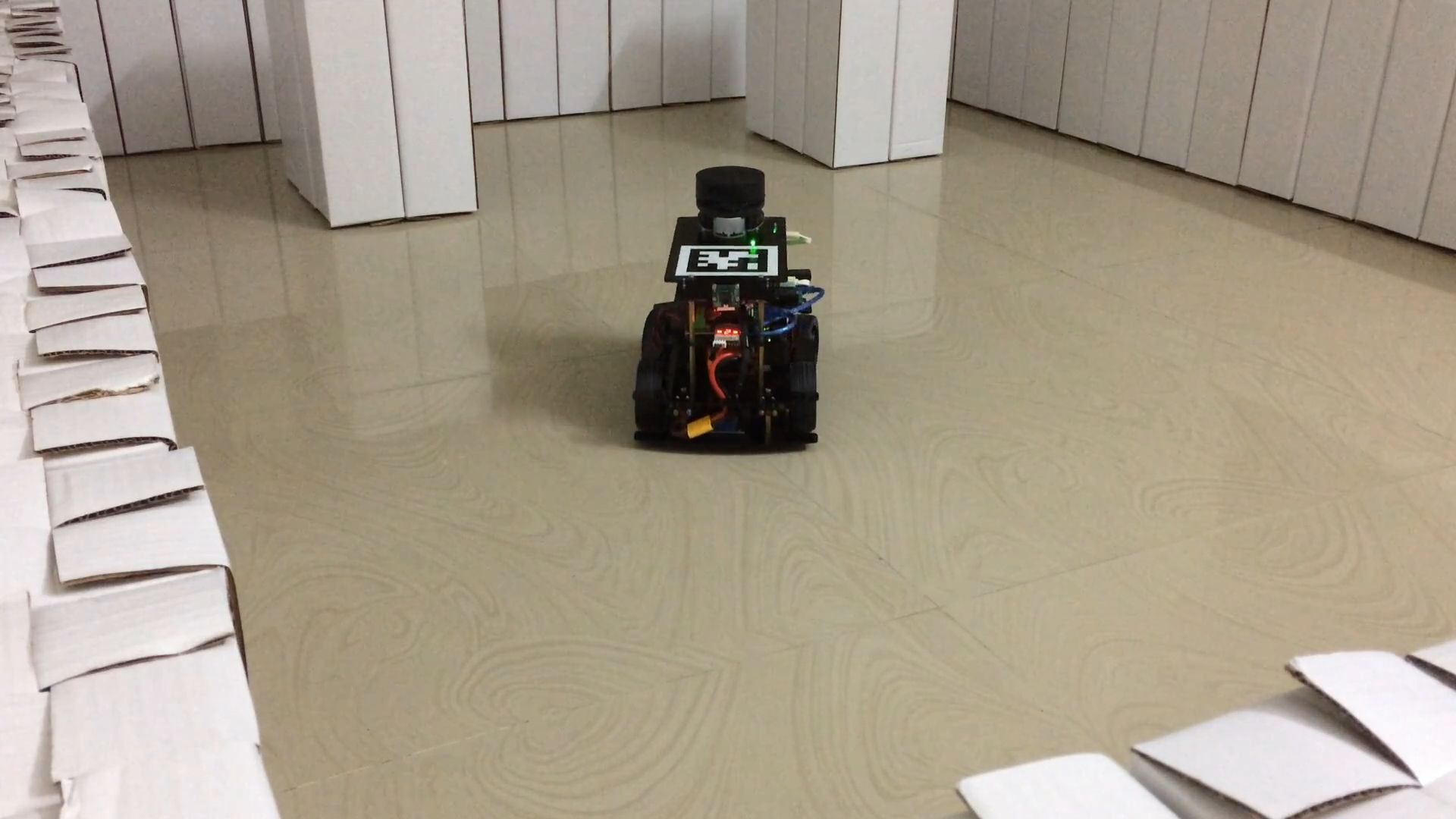}
         \caption{Parking School in reality.}
         \label{fig4b}
     \end{subfigure}
     \begin{subfigure}[b]{0.49\linewidth}
         \centering
         \includegraphics[width=\linewidth]{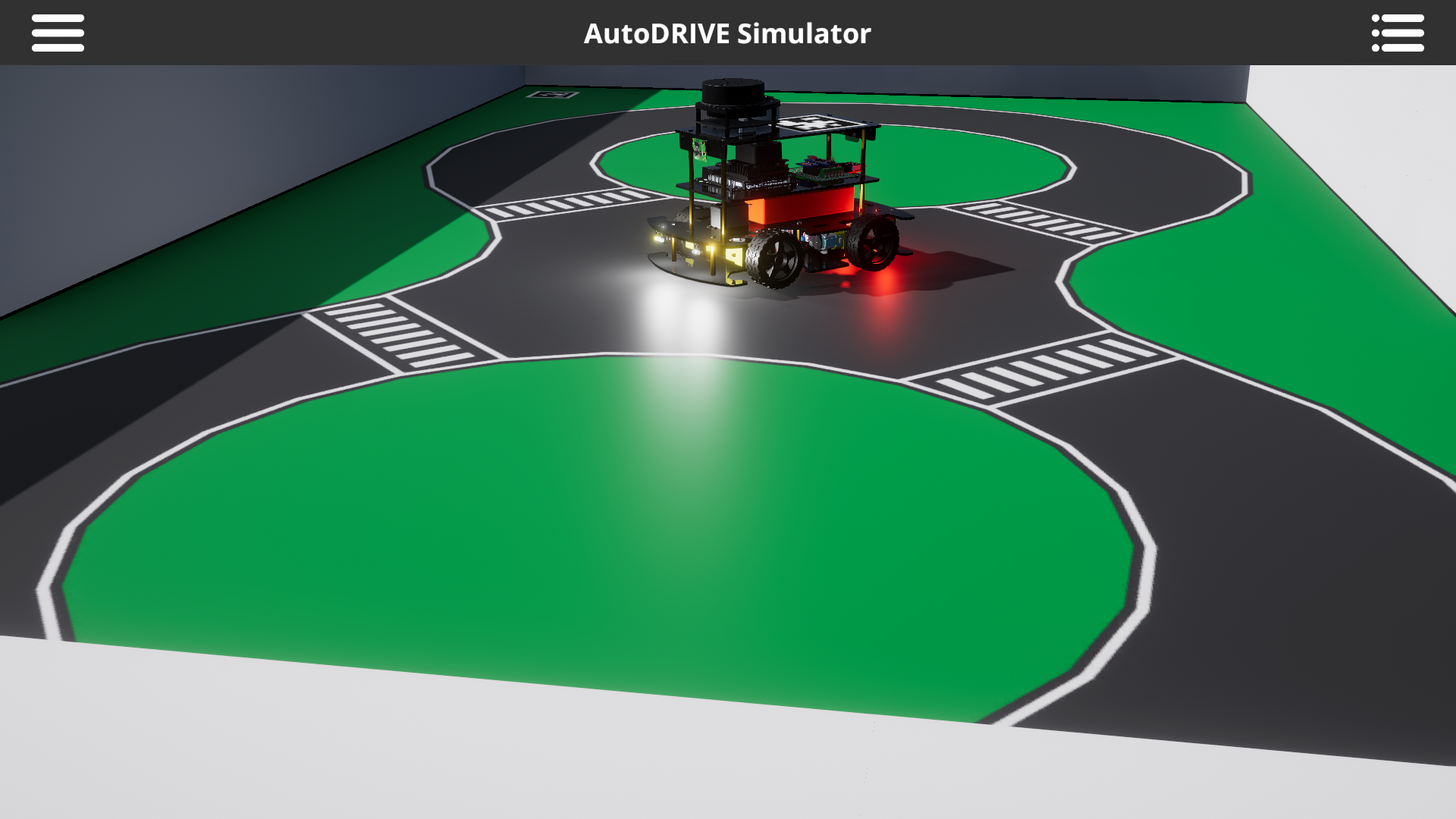}
         \caption{Driving School in simulation.}
         \label{fig4c}
     \end{subfigure}
     \begin{subfigure}[b]{0.49\linewidth}
         \centering
         \includegraphics[width=\linewidth]{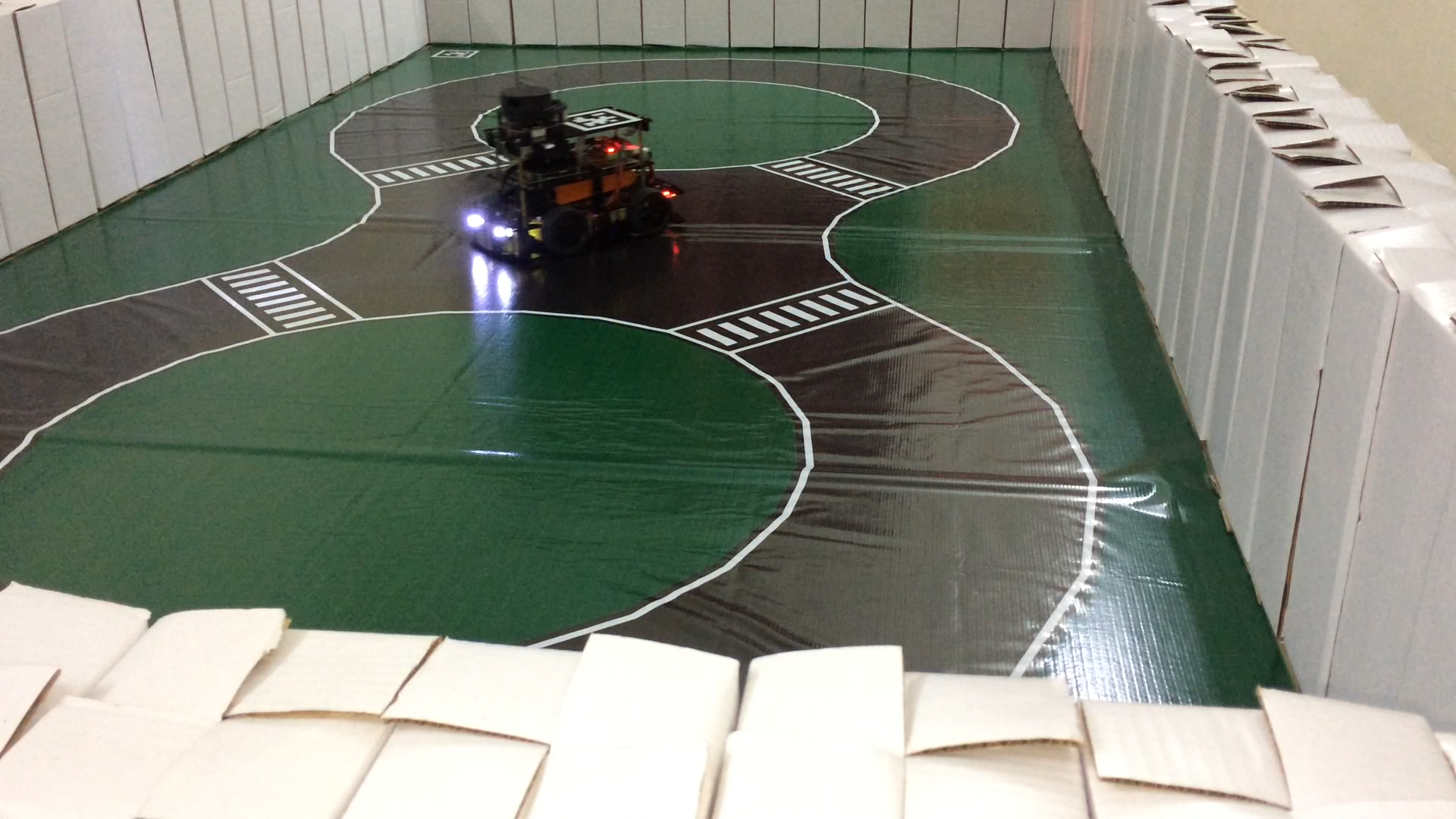}
         \caption{Driving School in reality.}
         \label{fig4d}
     \end{subfigure}
    \caption{Infrastructure setup in simulation and reality. Note the degree of dimensional and visual similarity between real and virtual worlds.}
    \label{fig4}
\end{figure}

\subsubsection{Environment Modules}
\label{Sub-Sub-Section: Environment Modules}

Environment modules include static terrain and road layouts as well as obstruction objects for rapidly designing custom scenarios. Apart from these, experts may also choose to design scaled real-world or imaginary scenarios using third-party tools, and import them into AutoDRIVE Ecosystem.

\subsubsection{Traffic Elements}
\label{Sub-Sub-Section: Traffic Elements}

Traffic signs and lights define traffic laws within a particular driving scenario, thereby governing the traffic flow. These modules support IoT and V2I communication technologies, and can be therefore integrated with AutoDRIVE Smart City Manager (SCM).

\subsubsection{Surveillance Elements}
\label{Sub-Sub-Section: Surveillance Elements}

AutoDRIVE features a surveillance element called AutoDRIVE Eye to view the scene from a bird's eye view. The said element is also integrated with AutoDRIVE SCM, and is capable of estimating vehicle's 2D pose within the map by detecting and tracking the AprilTag markers attached to each of them.

\subsubsection{Preconfigured Maps}
\label{Sub-Sub-Section: Preconfigured Maps}

This work focuses on two of the several preconfigured maps offered by AutoDRIVE Ecosystem. Parking School (refer Fig. \ref{fig4a}, \ref{fig4b}) is designed specifically for autonomous parking applications, wherein construction boxes define static obstacles and all the available free-space is drivable. On the other hand, Driving School (refer Fig. \ref{fig4c}, \ref{fig4d}) covers driving over straight roads, curved roads and crossing an intersection.

\subsection{Virtual Infrastructure}
\label{Section: Virtual Infrastructure}

At every time step, the simulator performs mesh-mesh collision/interference detection and accordingly computes the contact forces, frictional forces and momentum transfer, along with linear and angular drag acting over each of the rigid bodies (vehicle/infrastructure) present in the scene. This closely emulates the interactions among vehicle(s), infrastructure elements and the environment.


\section{Case Studies}
\label{Section: Case Studies}

This work showcases sim2real capability of AutoDRIVE Ecosystem through two different case-studies. Although this paper cannot furnish exhaustive details pertaining to either case study, we recommend interested readers to peruse this technical report \cite{AutoDRIVEReport2021}.

\begin{figure}[htpb]
     \centering
     \begin{subfigure}[b]{0.39\linewidth}
         \centering
         \includegraphics[height=1.02\linewidth]{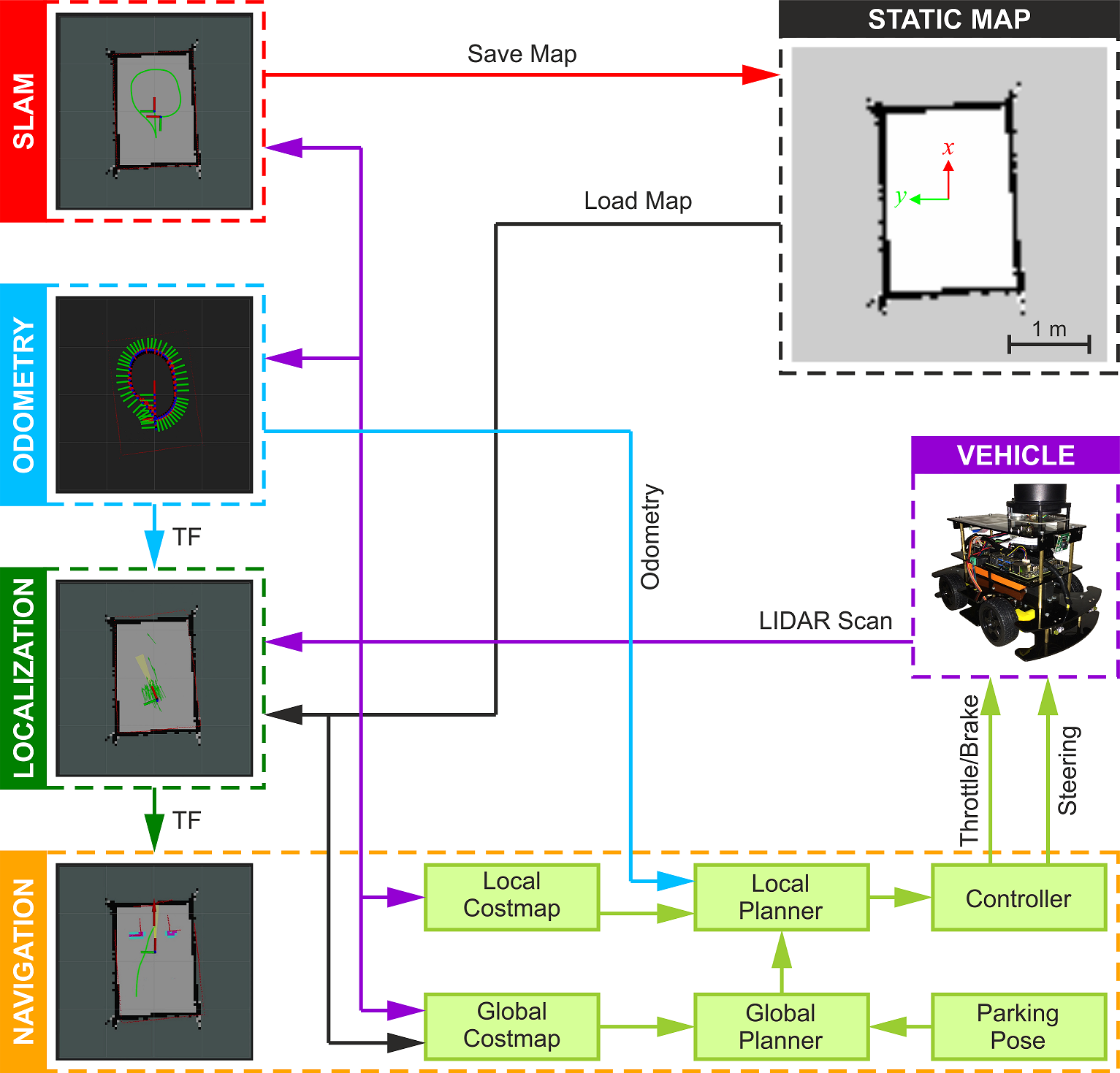}
         \caption{Autonomous parking.}
         \label{fig5a}
     \end{subfigure}
     \hfill
     \begin{subfigure}[b]{0.59\linewidth}
         \centering
         \includegraphics[height=0.68\linewidth]{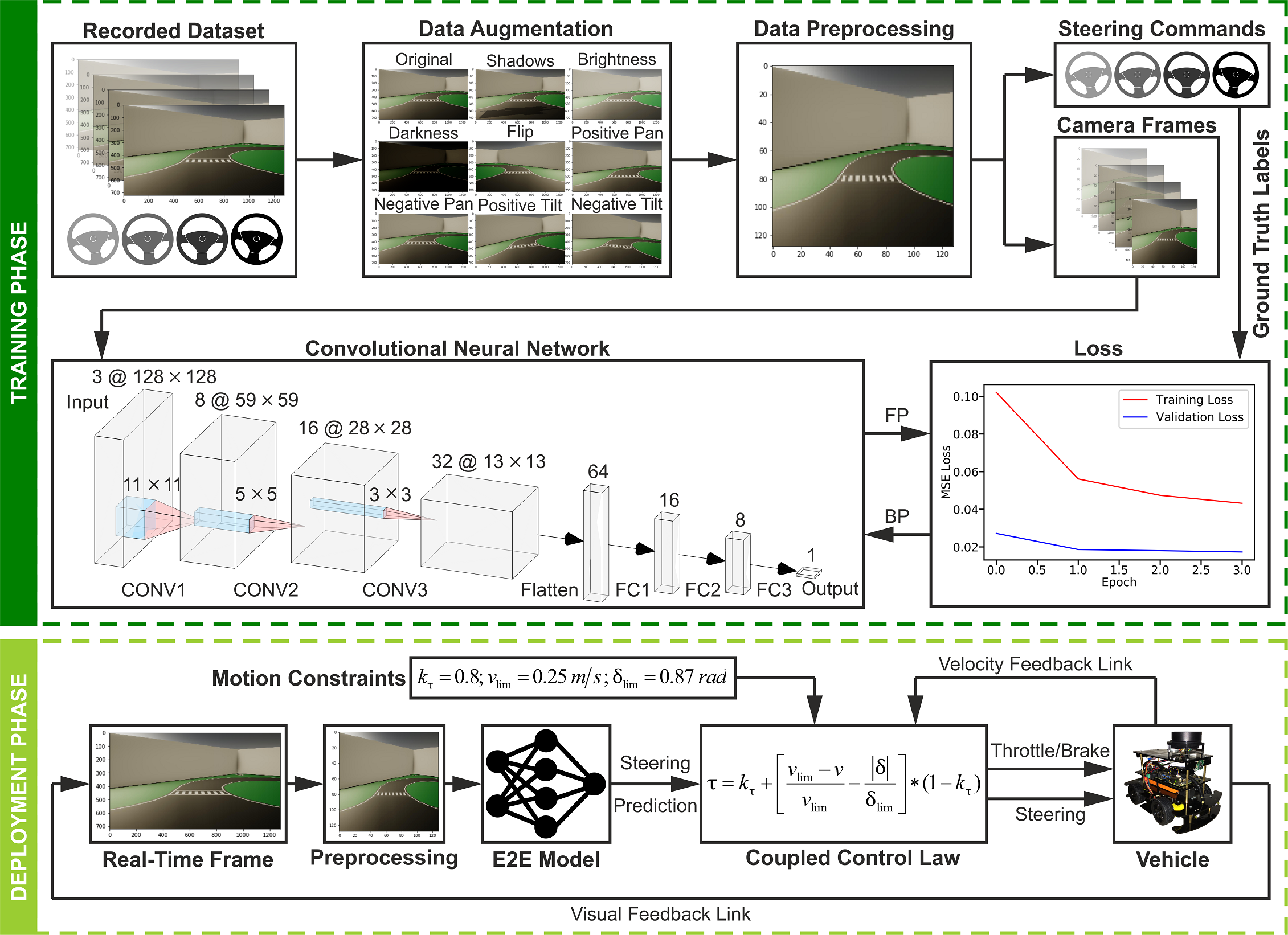}
         \caption{Behavioral cloning.}
         \label{fig5b}
     \end{subfigure}
    \caption{Architectures of the two presented case-studies.}
    \label{fig5}
\end{figure}

\subsection{Autonomous Parking}
\label{Sub-Section: Autonomous Parking}


This case study was implemented using the probabilistic robotics approach comprising 5 different stages. Firstly, the vehicle mapped its surroundings using the Hector SLAM algorithm \cite{HectorSLAM2011}. It then localized itself against this known static map using range-flow-based odometry \cite{RF2O2016} and an adaptive particle filter algorithm \cite{AMCL2001}. For autonomous navigation, the vehicle planned a feasible global path from its current pose to the parking pose using the A* algorithm \cite{AStar1968}. Simultaneously, it re-planned its local trajectory for dynamic collision avoidance, leveraging the timed-elastic-band approach \cite{TEBPlanner2017}. A proportional controller generated driving (throttle/brake) and steering commands to enable the vehicle to follow the local trajectory accurately.


\begin{figure}[htpb]
     \centering
     \begin{subfigure}[b]{0.49\linewidth}
         \centering
         \includegraphics[width=\linewidth]{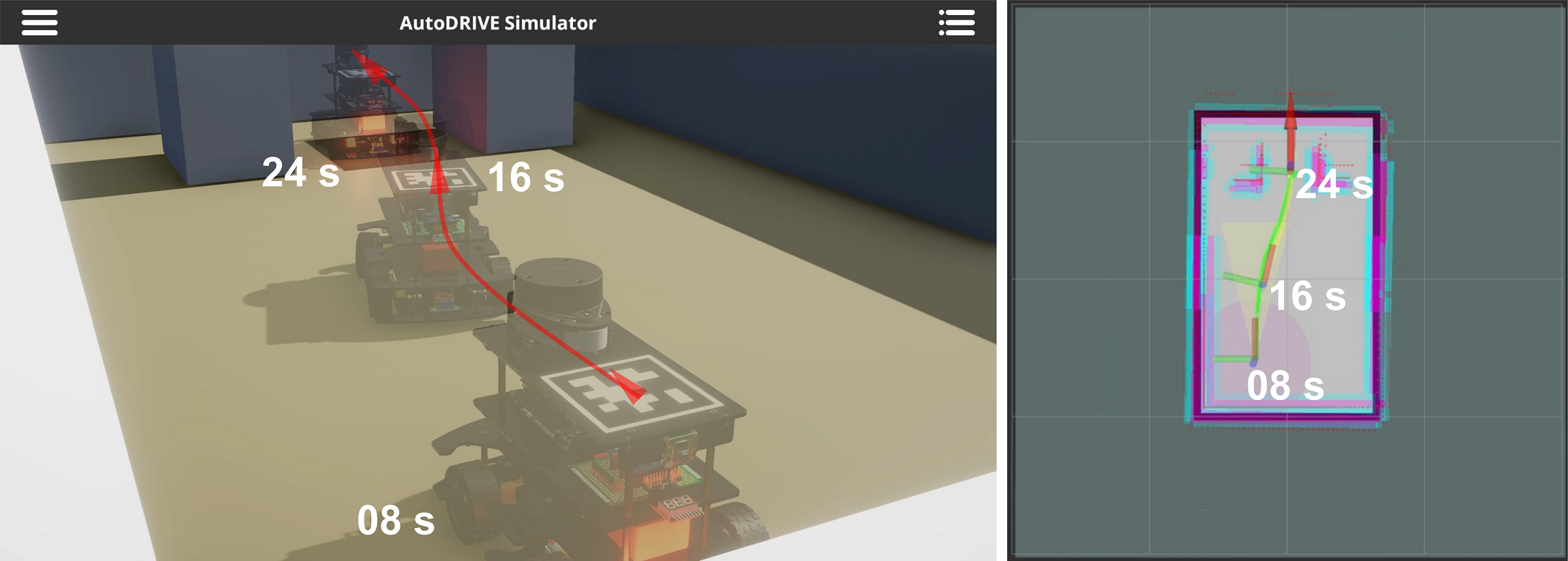}
         \caption{Virtual-world deployment.}
         \label{fig6a}
     \end{subfigure}
     \hfill
     \begin{subfigure}[b]{0.49\linewidth}
         \centering
         \includegraphics[width=\linewidth]{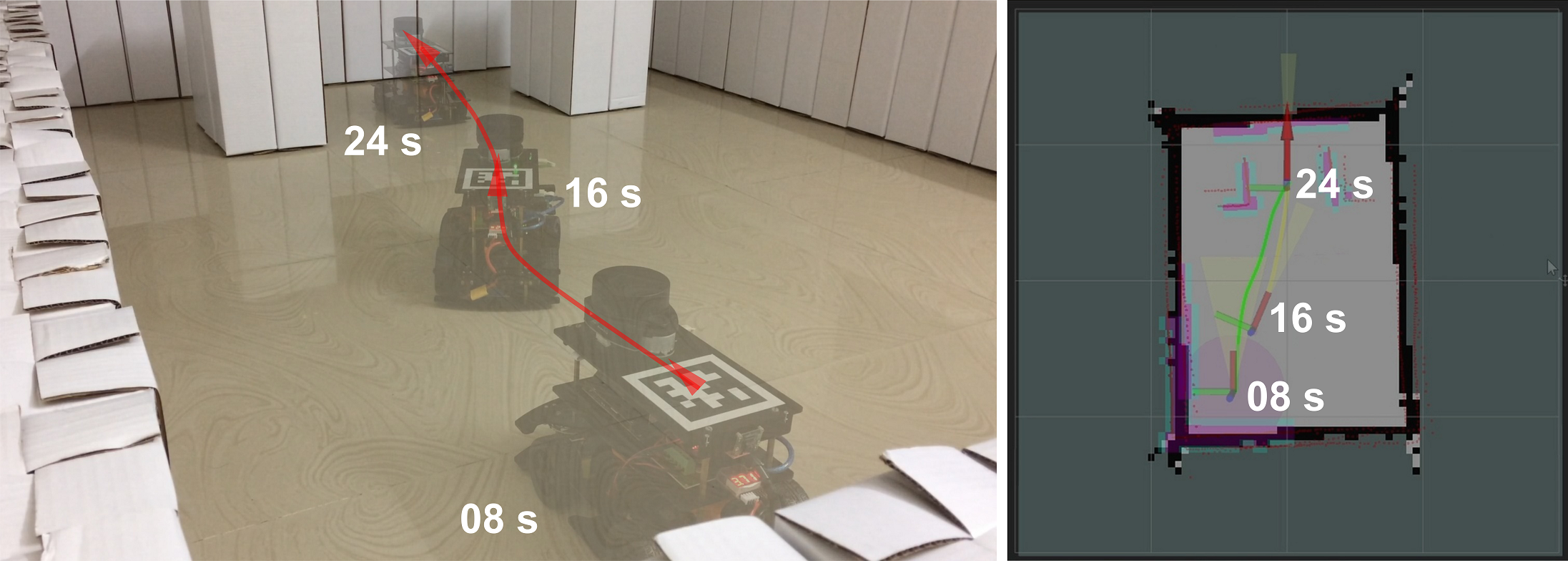}
         \caption{Real-world deployment.}
         \label{fig6b}
     \end{subfigure}
    \caption{Sim2real transition of autonomous parking algorithm. Video: \url{https://youtu.be/piCyvTM2dek}}
    \label{fig6}
\end{figure}



During simulation-based testing (refer Fig. \ref{fig6a}), parameter variations such as infusion of Gaussian noise in LIDAR measurements [$n_{lidar} \sim N\left ( 0, 0.025 \right )$ m] and actuator commands [$n_{drive} \sim N\left ( 0, 0.013 \right )$ m/s and $n_{steer} \sim N\left ( 0, 0.018 \right )$ rad/s] were introduced. Furthermore, perturbation of wall modules' poses [$n_{x,y} \sim N\left ( 0, 0.01 \right )$ m and $n_{\theta} \sim N\left ( 0, 0.087 \right )$ rad] as well as introduction of unmapped obstacles were performed. Going forward, the same pipeline was deployed on the real vehicle (refer Fig. \ref{fig6b}) using AutoDRIVE Testbed to validate seamless sim2real transfer. The pipeline performed flawlessly owing to realistic LIDAR and vehicle dynamics models as well as variability analysis during simulation.

\subsection{Behavioral Cloning}
\label{Sub-Section: Behavioral Cloning}

\begin{figure}[htpb]
     \centering
     \begin{subfigure}[b]{0.49\linewidth}
         \centering
         \includegraphics[width=\linewidth]{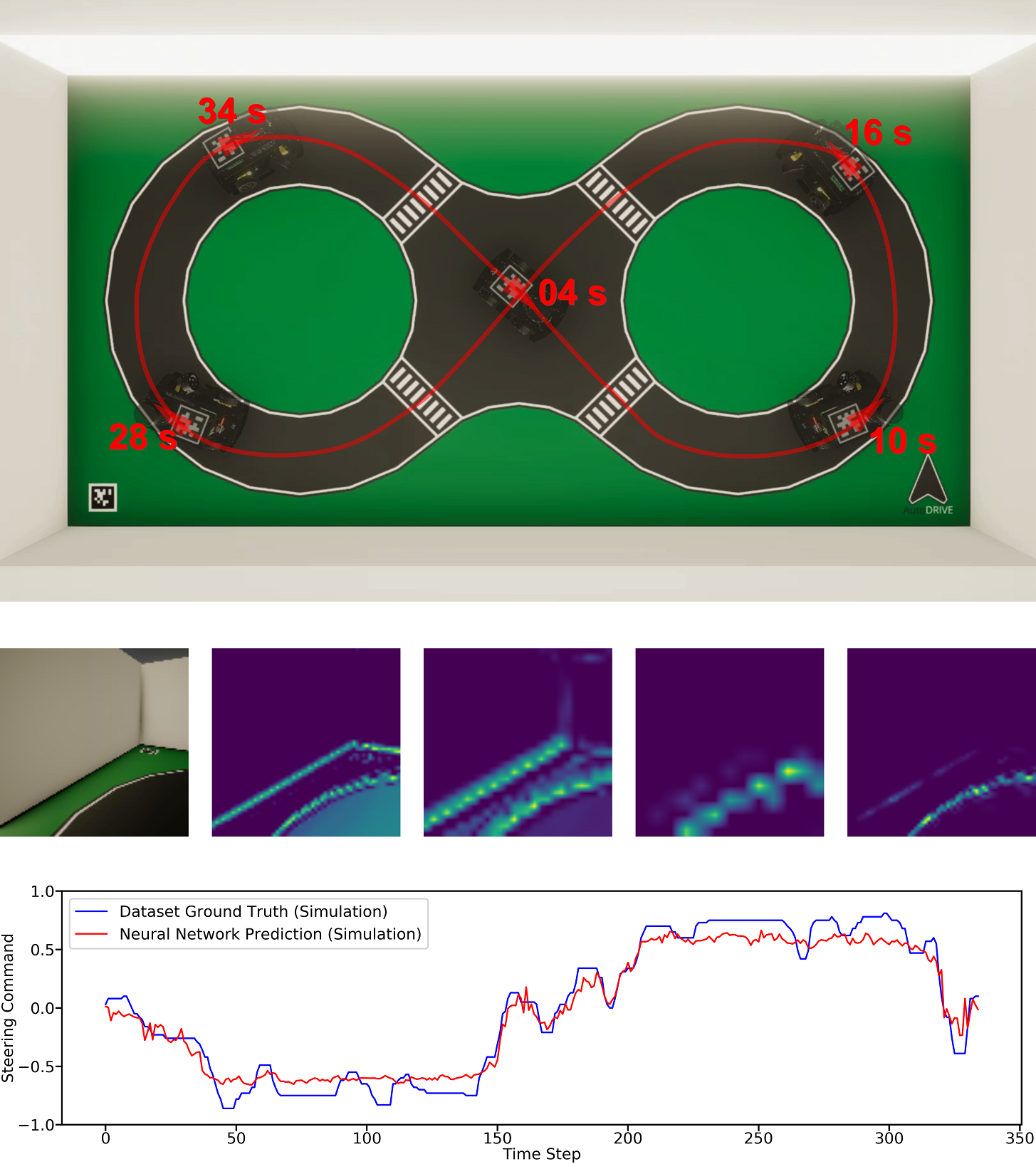}
         \caption{Virtual-world deployment.}
         \label{fig7a}
     \end{subfigure}
     \hfill
     \begin{subfigure}[b]{0.49\linewidth}
         \centering
         \includegraphics[width=\linewidth]{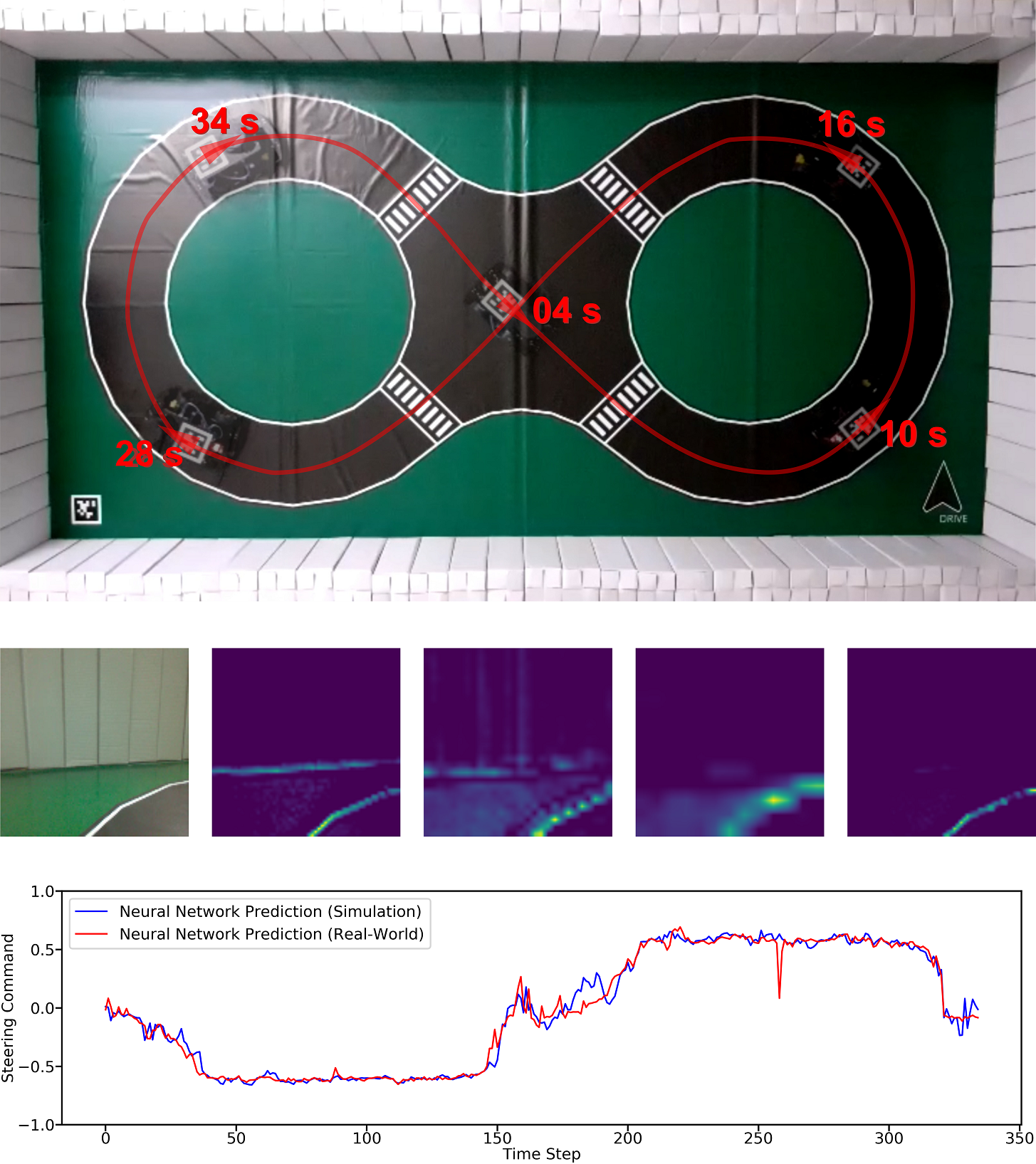}
         \caption{Real-world deployment.}
         \label{fig7b}
     \end{subfigure}
    \caption{Sim2real transition of behavioral cloning algorithm. Video: \url{https://youtu.be/rejpoogaXOE}}
    \label{fig7}
\end{figure}

This case study was based on \cite{RBC2021}, wherein the objective was to utilize a convolutional neural network (CNN) for cloning the end-to-end driving behavior of a human (refer to Fig. \ref{fig5b}). To achieve this, AutoDRIVE Simulator was employed to record 5-laps of temporally-coherent labeled manual driving data. This data was then balanced, augmented, and pre-processed using standard computer vision techniques to train a 6-layer deep CNN model for 4 epochs with a learning rate of 1e-3 using the Adam optimizer \cite{Kingma2014}. Following training, the model was deployed back into AutoDRIVE Simulator to evaluate its performance (refer to Fig. \ref{fig7a}).


Variability testing concerning light intensity and direction as well as vehicle's initial conditions and velocity limit was performed to ensure algorithm robustness. Further, the same CNN model was deployed onto AutoDRIVE Testbed for validating the sim2real transition of this vision-based algorithm (refer Fig. \ref{fig7b}). Despite subtle variations in the environmental conditions, the model generalized well and the vehicle was able to complete several laps along the track without major deviation and/or collision.


\section{Conclusion}
\label{Section: Conclusion}

In this paper, we presented AutoDRIVE, a publicly accessible digital twin ecosystem for CAVs developed with an aim of tightly integrating reality and simulation into a unified toolchain, without compromising on the comprehensiveness, flexibility and accessibility required for prototyping and validating autonomy solutions. This work focused on bridging the autonomy-oriented sim2real gap using the proposed ecosystem. Furthermore, we extensively discussed the modeling and simulation aspects of the ecosystem and substantiated its efficacy by demonstrating the successful transition of two candidate autonomy algorithms including autonomous parking and behavioral cloning from simulation to reality to help support our claims. Further research will delve into the investigation of handling real and virtual world uncertainties, formulating qualitative/quantitative evaluation metrics and benchmarks, as well as improving the robustness and generalization of sim2real frameworks for autonomous vehicles. 


\balance
\bibliography{ifacconf}

\begin{thebibliography}{29}
\providecommand{\natexlab}[1]{#1}
\providecommand{\url}[1]{\texttt{#1}}
\providecommand{\urlprefix}{URL }
\expandafter\ifx\csname urlstyle\endcsname\relax
  \providecommand{\doi}[1]{doi:\discretionary{}{}{}#1}\else
  \providecommand{\doi}{doi:\discretionary{}{}{}\begingroup
  \urlstyle{rm}\Url}\fi

\bibitem[{{Baidu Inc.}(2021)}]{ApolloGameSim2021}
{Baidu Inc.} (2021).
\newblock {Apollo Game Engine Based Simulator}.

\bibitem[{Brockman et~al.(2016)Brockman, Cheung, Pettersson, Schneider,
  Schulman, Tang, and Zaremba}]{OpenAIGym2016}
Brockman, G., Cheung, V., Pettersson, L., Schneider, J., Schulman, J., Tang,
  J., and Zaremba, W. (2016).
\newblock {OpenAI Gym}.

\bibitem[{{Donkey Community}(2021)}]{DonkeyCar2021}
{Donkey Community} (2021).
\newblock {An Open-Source DIY Self-Driving Platform for Small-Scale Cars}.

\bibitem[{Dosovitskiy et~al.(2017)Dosovitskiy, Ros, Codevilla, Lopez, and
  Koltun}]{CARLA2017}
Dosovitskiy, A., Ros, G., Codevilla, F., Lopez, A., and Koltun, V. (2017).
\newblock {CARLA: An Open Urban Driving Simulator}.
\newblock In S.~Levine, V.~Vanhoucke, and K.~Goldberg (eds.), \emph{Proceedings
  of the 1st Annual Conference on Robot Learning}, volume~78 of
  \emph{Proceedings of Machine Learning Research}, 1--16. PMLR.

\bibitem[{Fox(2001)}]{AMCL2001}
Fox, D. (2001).
\newblock {KLD-Sampling: Adaptive Particle Filters}.
\newblock In T.~Dietterich, S.~Becker, and Z.~Ghahramani (eds.), \emph{Advances
  in Neural Information Processing Systems}, volume~14. MIT Press.

\bibitem[{Goldfain et~al.(2019)Goldfain, Drews, You, Barulic, Velev, Tsiotras,
  and Rehg}]{AutoRally2021}
Goldfain, B., Drews, P., You, C., Barulic, M., Velev, O., Tsiotras, P., and
  Rehg, J.M. (2019).
\newblock {AutoRally: An Open Platform for Aggressive Autonomous Driving}.
\newblock \emph{IEEE Control Systems Magazine}, 39(1), 26--55.
\newblock \doi{10.1109/MCS.2018.2876958}.

\bibitem[{Hart et~al.(1968)Hart, Nilsson, and Raphael}]{AStar1968}
Hart, P.E., Nilsson, N.J., and Raphael, B. (1968).
\newblock {A Formal Basis for the Heuristic Determination of Minimum Cost
  Paths}.
\newblock \emph{IEEE Transactions on Systems Science and Cybernetics}, 4(2),
  100--107.
\newblock \doi{10.1109/TSSC.1968.300136}.

\bibitem[{Hershberger et~al.(2021)Hershberger, Gossow, and Faust}]{RViz2021}
Hershberger, D., Gossow, D., and Faust, J. (2021).
\newblock {RViz: 3D Visualization Tool for ROS}.

\bibitem[{{HyphaROS Workshop}(2021)}]{HyphaROS-Racecar2021}
{HyphaROS Workshop} (2021).
\newblock {HyphaROS Racecar}.

\bibitem[{Jaimez et~al.(2016)Jaimez, Monroy, and Gonzalez-Jimenez}]{RF2O2016}
Jaimez, M., Monroy, J.G., and Gonzalez-Jimenez, J. (2016).
\newblock {Planar Odometry from a Radial Laser Scanner. A Range Flow-based
  Approach}.
\newblock In \emph{2016 IEEE International Conference on Robotics and
  Automation (ICRA)}, 4479--4485.
\newblock \doi{10.1109/ICRA.2016.7487647}.

\bibitem[{Karaman et~al.(2017)Karaman, Anders, Boulet, Connor, Gregson, Guerra,
  Guldner, Mohamoud, Plancher, Shin, and Vivilecchia}]{MIT-Racecar2017}
Karaman, S., Anders, A., Boulet, M., Connor, J., Gregson, K., Guerra, W.,
  Guldner, O., Mohamoud, M., Plancher, B., Shin, R., and Vivilecchia, J.
  (2017).
\newblock {Project-based, collaborative, algorithmic robotics for high school
  students: Programming self-driving race cars at MIT}.
\newblock In \emph{2017 IEEE Integrated STEM Education Conference (ISEC)},
  195--203.
\newblock \doi{10.1109/ISECon.2017.7910242}.

\bibitem[{Kingma and Ba(2014)}]{Kingma2014}
Kingma, D.P. and Ba, J. (2014).
\newblock {Adam: A Method for Stochastic Optimization}.
\newblock \doi{10.48550/ARXIV.1412.6980}.

\bibitem[{Koenig and Howard(2004)}]{Gazebo2004}
Koenig, N.P. and Howard, A. (2004).
\newblock {Design and use paradigms for Gazebo, an open-source multi-robot
  simulator}.
\newblock In \emph{2004 IEEE/RSJ International Conference on Intelligent Robots
  and Systems (IROS) (IEEE Cat. No.04CH37566)}, volume~3, 2149--2154.
\newblock \doi{10.1109/IROS.2004.1389727}.

\bibitem[{Kohlbrecher et~al.(2011)Kohlbrecher, von Stryk, Meyer, and
  Klingauf}]{HectorSLAM2011}
Kohlbrecher, S., von Stryk, O., Meyer, J., and Klingauf, U. (2011).
\newblock {A Flexible and Scalable SLAM System with Full 3D Motion Estimation}.
\newblock In \emph{2011 IEEE International Symposium on Safety, Security, and
  Rescue Robotics}, 155--160.
\newblock \doi{10.1109/SSRR.2011.6106777}.

\bibitem[{O'Kelly et~al.(2019)O'Kelly, Sukhil, Abbas, Harkins, Kao, Pant,
  Mangharam, Agarwal, Behl, Burgio, and Bertogna}]{F1TENTH2019}
O'Kelly, M., Sukhil, V., Abbas, H., Harkins, J., Kao, C., Pant, Y.V.,
  Mangharam, R., Agarwal, D., Behl, M., Burgio, P., and Bertogna, M. (2019).
\newblock {F1/10: An Open-Source Autonomous Cyber-Physical Platform}.

\bibitem[{Paull et~al.(2017)Paull, Tani, Ahn, Alonso-Mora, Carlone, Cap, Chen,
  Choi, Dusek, Fang, Hoehener, Liu, Novitzky, Okuyama, Pazis, Rosman,
  Varricchio, Wang, Yershov, Zhao, Benjamin, Carr, Zuber, Karaman, Frazzoli,
  Del~Vecchio, Rus, How, Leonard, and Censi}]{Duckietown2017}
Paull, L., Tani, J., Ahn, H., Alonso-Mora, J., Carlone, L., Cap, M., Chen,
  Y.F., Choi, C., Dusek, J., Fang, Y., Hoehener, D., Liu, S.Y., Novitzky, M.,
  Okuyama, I.F., Pazis, J., Rosman, G., Varricchio, V., Wang, H.C., Yershov,
  D., Zhao, H., Benjamin, M., Carr, C., Zuber, M., Karaman, S., Frazzoli, E.,
  Del~Vecchio, D., Rus, D., How, J., Leonard, J., and Censi, A. (2017).
\newblock {Duckietown: An Open, Inexpensive and Flexible Platform for Autonomy
  Education and Research}.
\newblock In \emph{2017 IEEE International Conference on Robotics and
  Automation (ICRA)}, 1497--1504.
\newblock \doi{10.1109/ICRA.2017.7989179}.

\bibitem[{Quigley et~al.(2009)Quigley, Conley, Gerkey, Faust, Foote, Leibs,
  Wheeler, and Ng}]{ROS2009}
Quigley, M., Conley, K., Gerkey, B., Faust, J., Foote, T., Leibs, J., Wheeler,
  R., and Ng, A. (2009).
\newblock {ROS: an open-source Robot Operating System}.
\newblock In \emph{ICRA Workshop on Open Source Software}, volume~3.

\bibitem[{{Robotis Inc.}(2021)}]{Turtlebot2021}
{Robotis Inc.} (2021).
\newblock {TurtleBot3}.

\bibitem[{Rong et~al.(2020)Rong, Shin, Tabatabaee, Lu, Lemke, Možeiko, Boise,
  Uhm, Gerow, Mehta, Agafonov, Kim, Sterner, Ushiroda, Reyes, Zelenkovsky, and
  Kim}]{LGSVLSimulator2020}
Rong, G., Shin, B.H., Tabatabaee, H., Lu, Q., Lemke, S., Možeiko, M., Boise,
  E., Uhm, G., Gerow, M., Mehta, S., Agafonov, E., Kim, T.H., Sterner, E.,
  Ushiroda, K., Reyes, M., Zelenkovsky, D., and Kim, S. (2020).
\newblock {LGSVL Simulator: A High Fidelity Simulator for Autonomous Driving}.
\newblock In \emph{2020 IEEE 23rd International Conference on Intelligent
  Transportation Systems (ITSC)}, 1--6.
\newblock \doi{10.1109/ITSC45102.2020.9294422}.

\bibitem[{Rösmann et~al.(2017)Rösmann, Hoffmann, and
  Bertram}]{TEBPlanner2017}
Rösmann, C., Hoffmann, F., and Bertram, T. (2017).
\newblock {Kinodynamic Trajectory Optimization and Control for Car-Like
  Robots}.
\newblock In \emph{2017 IEEE/RSJ International Conference on Intelligent Robots
  and Systems (IROS)}, 5681--5686.
\newblock \doi{10.1109/IROS.2017.8206458}.

\bibitem[{Samak et~al.(2021{\natexlab{a}})Samak, Samak, and
  Kandhasamy}]{AutoRACE2021}
Samak, C.V., Samak, T.V., and Kandhasamy, S. (2021{\natexlab{a}}).
\newblock {Autonomous Racing using a Hybrid Imitation-Reinforcement Learning
  Architecture}.
\newblock \doi{10.48550/ARXIV.2110.05437}.

\bibitem[{Samak et~al.(2023)Samak, Samak, Kandhasamy, Krovi, and
  Xie}]{AutoDRIVEEcosystem2023}
Samak, T., Samak, C., Kandhasamy, S., Krovi, V., and Xie, M. (2023).
\newblock {AutoDRIVE: A Comprehensive, Flexible and Integrated Digital Twin
  Ecosystem for Autonomous Driving Research \& Education}.
\newblock \emph{Robotics}, 12(3).
\newblock \doi{10.3390/robotics12030077}.

\bibitem[{Samak and Samak(2022{\natexlab{a}})}]{AutoDRIVEReport2021}
Samak, T.V. and Samak, C.V. (2022{\natexlab{a}}).
\newblock {AutoDRIVE - Technical Report}.
\newblock \doi{10.48550/ARXIV.2211.08475}.

\bibitem[{Samak and Samak(2022{\natexlab{b}})}]{AutoDRIVESimulatorReport2020}
Samak, T.V. and Samak, C.V. (2022{\natexlab{b}}).
\newblock {AutoDRIVE Simulator - Technical Report}.
\newblock \doi{10.48550/ARXIV.2211.07022}.

\bibitem[{Samak et~al.(2021{\natexlab{b}})Samak, Samak, and
  Kandhasamy}]{RBC2021}
Samak, T.V., Samak, C.V., and Kandhasamy, S. (2021{\natexlab{b}}).
\newblock {Robust Behavioral Cloning for Autonomous Vehicles Using End-to-End
  Imitation Learning}.
\newblock \emph{SAE International Journal of Connected and Automated Vehicles},
  4(3), 279--295.
\newblock \doi{https://doi.org/10.4271/12-04-03-0023}.

\bibitem[{Samak et~al.(2021{\natexlab{c}})Samak, Samak, and
  Xie}]{AutoDRIVESimulator2021}
Samak, T.V., Samak, C.V., and Xie, M. (2021{\natexlab{c}}).
\newblock {AutoDRIVE Simulator: A Simulator for Scaled Autonomous Vehicle
  Research and Education}.
\newblock In \emph{2021 2nd International Conference on Control, Robotics and
  Intelligent System}, CCRIS'21, 1–5. Association for Computing Machinery,
  New York, NY, USA.
\newblock \doi{10.1145/3483845.3483846}.

\bibitem[{Shah et~al.(2018)Shah, Dey, Lovett, and Kapoor}]{AirSim2018}
Shah, S., Dey, D., Lovett, C., and Kapoor, A. (2018).
\newblock {AirSim: High-Fidelity Visual and Physical Simulation for Autonomous
  Vehicles}.
\newblock In M.~Hutter and R.~Siegwart (eds.), \emph{Field and Service
  Robotics}, 621--635. Springer International Publishing, Cham.

\bibitem[{{Voyage}(2021)}]{Deepdrive2021}
{Voyage} (2021).
\newblock {Deepdrive}.

\bibitem[{Wymann et~al.(2021)Wymann, Espié, Guionneau, Dimitrakakis, Coulom,
  and Sumner}]{TORCS2021}
Wymann, B., Espié, E., Guionneau, C., Dimitrakakis, C., Coulom, R., and
  Sumner, A. (2021).
\newblock {TORCS, The Open Racing Car Simulator}.

\end{thebibliography}


\end{document}